\documentclass[journal]{IEEEtran}
\ifCLASSINFOpdf
\else
\fi
\usepackage{cite}
\usepackage{graphicx}
\usepackage{amsmath}
\usepackage{amsfonts,amssymb}
\usepackage{mathrsfs}
\usepackage{array}
\usepackage{multirow}
\usepackage{longtable}
\usepackage{rotating}

\begin{document}
\title{Common Spatial Generative Adversarial Networks based EEG Data Augmentation for Cross-Subject Brain-Computer Interface}

\author{Yonghao Song, Lie Yang, Xueyu Jia,
        Longhan~Xie,~\IEEEmembership{Member,~IEEE} 
\thanks{(corresponding author: Longhan Xie)}
\thanks{Yonghao Song, Lie Yang, Xueyu Jia and Longhan Xie are with the Shien-Ming Wu School of Intelligent Engineering, South China University of Technology, Guangzhou 510460, China (e-mail: eeyhsong@gmail.com, xielonghan@gmail.com). }}

\markboth{}%
{Shell \MakeLowercase{\textit{et al.}}: Bare Demo of IEEEtran.cls for IEEE Journals}

\maketitle

\begin{abstract}
The cross-subject application of EEG-based brain-computer interface (BCI) has always been limited by large individual difference and complex characteristics that are difficult to perceive. Therefore, it takes a long time to collect the training data of each user for calibration. Even transfer learning method pre-training with amounts of subject-independent data cannot decode different EEG signal categories without enough subject-specific data. Hence, we proposed a cross-subject EEG classification framework with a generative adversarial networks (GANs) based method named common spatial GAN (CS-GAN), which used adversarial training between a generator and a discriminator to obtain high-quality data for augmentation. A particular module in the discriminator was employed to maintain the spatial features of the EEG signals and increase the difference between different categories, with two losses for further enhancement. Through adaptive training with sufficient augmentation data, our cross-subject classification accuracy yielded a significant improvement of 15.85\% than leave-one subject-out (LOO) test and 8.57\% than just adapting 100 original samples on the dataset 2a of BCI competition IV. Moreover, We designed a convolutional neural networks (CNNs) based classification method as a benchmark with a similar spatial enhancement idea, which achieved remarkable results to classify motor imagery EEG data. In summary, our framework provides a promising way to deal with the cross-subject problem and promote the practical application of BCI.
\end{abstract}

\begin{IEEEkeywords}
Electroencephalograph (EEG), generative adversarial networks (GANs), data augmentation, brain-computer interface (BCI), motor imagery (MI), cross-subject.
\end{IEEEkeywords}

\IEEEpeerreviewmaketitle

\section{Introduction}
%
%
%
%
\IEEEPARstart{R}{esearchers} have been trying to decode the information in the brain for many years. One of the commonly used in the field is electroencephalograph (EEG), a non-invasive monitoring method to record brain neurons’ electrical activity with some electrodes on the scalp \cite{[1]}. Because of the stable temporal and spatial resolution, EEG has achieved good performance in diagnosing some diseases such as epilepsy, insomnia and Alzheimer’s \cite{[2],[3],[4]}, and also brings great possibilities in brain-computer interface (BCI) \cite{[5]}.

BCI is a technology that aims to establish pathways between the brain and external devices, commonly used in robot control, entertainment and rehabilitation \cite{[6],[7]}. Motor imagery is decoded to aid paralyzed people with physical therapy, which has proven to be beneficial for rehabilitation after a stroke or spinal cord injury. \cite{[8],[9],[10]}. 

With the improvement of classification methods, the decoding of motor intention is becoming increasingly accurate. However, most research is to train classifiers on a single subject instead of putting together the data of different subjects \cite{[11],[12],[13]}, which means that sufficient training data have to be collected for a new user. It is obviously unfeasible in many scenarios, especially for some patients. 

The classification problem called cross-subject is mainly restricted for two reasons. One reason is that EEG signals are non-stationary with large individual differences caused by different physiological characteristics \cite{[14]}. Like in facial expression recognition research, the effect of classification performance is severely affected by identity information \cite{[15]}. Furthermore, the limited amount of subject-specific EEG data is insufficient to support some good methods such as convolutional neural networks (CNNs) perceiving individual features related to different categories \cite{[16]}. Subsequently, transfer learning has been implemented to cross-subject problems by pre-training with subject-independent data firstly, but still not competent unless there is enough data of target subject for fine-tuning \cite{[17]}. Naturally, researchers thought of artificially generating EEG data to resemble and augment the training set. Methods such as adding Gaussian noise and segmentation have been used and obtained a certain extent of improvement \cite{[18],[19]}. The redundant noise or the loss of information does not meet our needs yet. With an emphasis on data generation, the generative models maybe the potential solution to this dilemma. In particular, generative adversarial networks (GANs) have attracted great attention in computer vision due to the excellent artificial image generation capabilities \cite{[20]}.

Another reason limiting the cross-subject problem is the low signal-to-noise ratio of the EEG signals, which are easily interfered by various factors such as impedance and muscle artifacts. It is also prone to draw into a mass of irrelevant information when the subjects are not concentrated. This obstacle has not been handled well just using original temporal features with end-to-end machine learning methods. Therefore, some work added feature extraction processing before the classifier and confirmed that it makes some sense. This processing could be summarized into two types. One is to calculate different features of the signal, such as autoregressive model to obtain time-series features \cite{[21]}, power spectrum to obtain spectral features and wavelet transforms to obtain time-frequency features \cite{[22],[23]}. The other is to project the original signal into a subspace while obtaining spatial features for better classification, such as optimal spatial filter and xDAWN algorithm \cite{[24],[25]}. 

Although the existing works have had some improvements in EEG decoding, few of them really deal with the cross-subject problem. In this article, therefore, we propose a GAN framework to augment multi-channel EEG data with motor intention, and preserve the spatial features in addition to temporal features, while enhancing the discrimination between different categories of generated signals in a subspace. To verify the potential of this framework named common spatial GAN (CS-GAN), a well-designed CNN model that emphasizes spatial enhancement for multi-task was given firstly as a benchmark classification method and has achieved the state-of-the-art single-subject results on BCI competition IV Datasets 2a. Then, a minimal amount of subject-specific data was employed by CS-GAN to obtain generated data, which was used to adaptively augment the training set composed of subject-independent data. The results showed that the augmentation with CS-GAN framework yielded a more significant improvement on cross-subject classification, compared to other existing augmentation approaches. Besides, detailed analysis has been conducted on the temporal, spatial and frequency features of the generated EEG signal to show the authenticity. Overall, the CS-GAN assessments from different perspectives demonstrate that it is a promising method to promote cross-subject problem and improve BCI systems' usability. 

The main contributions of this work can be summarized as follows.  

1) We propose the CS-GAN framework for EEG data augmentation. Not only are EEG signals generated with good temporal pattern, but their spatial characteristics are also preserved well with the categories difference increased. 

2) With the data augmenting by CS-GAN, we improve the EEG classification performance under the cross-subject condition to a significant measure, which helps reduce the calibration time of BCI systems.

3) We also design a CNN model as a benchmark for motor intention classification with EEG signals projected to a new space, where the spatial difference is enhanced and the temporal information is retained. Remarkable performance on BCI competition IV Datasets 2a has been obtained.

4) Besides, we provide a feasible attempt to use GANs-based framework for data augmentation with just a few data.

\section{Related works}
\subsection{GANs and GANs for EEG}
GANs are a machine learning strategy inspired by game theory. Goodfellow \textit{et al.} proposed this network consisting of a generator and a discriminator for image generation firstly in 2014 \cite{[20]}. The generator is used to generate fake data similar to the real data from random series by estimating the original data distribution, and the discriminator is to discriminate whether the generated data are real or fake. After multiple rounds of adversarial training, the two modules gradually reach equilibrium, when the generator could create very real data that the discriminator cannot distinguish.

Research has emerged to solve several limitations of the first version of GAN. Conditional GANs proposed by Mirza \textit{et al.} \cite{[26]} and auxiliary classifier GANs proposed by Odena \textit{et al.} \cite{[27]} introduced category information as a prior condition into the generator and discriminator to generate samples of the specified category. Radford \textit{et al.} presented deep convolutional GANs to build a bridge between supervised and unsupervised learning by adding convolutional structures. Features were easy to capture, producing more delicate images \cite{[28]}. It is worth noting that Earth-Mover divergence was given by Arjovsky \textit{et al.} in Wasserstein GANs (WGANs) \cite{[29]}, which and the gradient penalty \cite{[30]} greatly improve the stability of GANs training. Some exciting attempts have also been put forward, such as image-to-image translation \cite{[31]} and image inpainting \cite{[32]}.

The amazing ability of GANs to generate artificial data quickly grabbed BCI researchers. Hartmann \textit{et al.} proposed the EEG-GAN to generate single-channel EEG signals with very good visual inspection \cite{[33]}. Roy \textit{et al.} employed long short-term memory networks in generator and discriminator and obtained motor imagery EEG signals with the same dynamic and time-frequency characteristics as the original signals \cite{[34]}. After being convinced that GANs have the potential to generate EEG, non-stationary time series, different applications have been tried, such as up-sampling EEG spatial resolution \cite{[35]}, session-invariant representation learning \cite{[36]} and data augmentation. Luo \textit{et al.} implemented a conditional Wasserstein GAN to generate power spectral density and differential entropy of EEG signals for enhancing EEG-based emotion recognition \cite{[37]}. Zhang \textit{et al.} utilized a conditional deep convolutional GAN to augment data after wavelet transform was applied \cite{[38]}. In addition to generating EEG features, researchers have also tried to generate unprocessed EEG signals for broader purposes. Aznan \textit{et al.} compared three generative models for EEG data generation, which was proven to be beneficial in EEG classification used in online control of humanoid robots \cite{[39]}. Fahimi \textit{et al.} extracted a feature vector from the target subject's data as a condition into GANs and obtained the multi-channel EEG signals that inherited the specific characteristics of the subject \cite{[40]}. This research shows a significant improvement to classify motor imagery EEG. Does it also indicate that the possibility of GANs is not limited to generating more EEG signals? 

\subsection{EEG Feature Extraction and Cross-subject Task}
Various EEG feature extraction methods have been tried to decode EEG better. Zhao \textit{et al.} arranged the channels according to the spatial distribution at each time point to obtain 3D data with more spatial information \cite{[41]}. Fan \textit{et al.} gained the spectral graph theoretic features to quantify the temporal synchronization for detecting abnormal patterns of epileptic seizures with EEG \cite{[42]}. Fast Fourier transform (FFT) and continuous wavelet transform (CWT) was applied by Durongbhan \textit{et al.} to extract the frequency and time-frequency features and construct a framework that uses EEG to classify Alzheimer's participants and the healthy \cite{[43]}. Spatial filter and some of its extensions, which find an optimal spatial filter to maximize the difference between two categories, have also achieved promising results. The filter bank common spatial filter (FBCSP) proposed by Ang \textit{et al.} decomposes EEG data into nine frequency bands, gets the spatial features respectively, and then selects the features with mutual information-based algorithm for better classification \cite{[44]}. Some research employed end-to-end models that embedded a feature extractor and classifier into a deep neural network to conduct joint optimization. Three convolutional blocks of a CNN were designed by Gao \textit{et al.} to get spatial-temporal features of EEG \cite{[45]}, similar to the way of Li \textit{et al.} \cite{[46]}.

Feature-oriented methods are often used for cross-subject tasks. Handiru \textit{et al.} presented a channel selection method to find the most relevant common features of motor imagery \cite{[47]}. Gupta \textit{et al.} decomposed the EEG signal into subbands with flexible analytic wavelet transform and applied information potential to extract the features for cross-subject emotion recognition \cite{[48]}. A popular technique is recently used for cross-subject problems termed as transfer learning or domain adaption, which extracts important information or pre-trains classifiers with the training data of source domain, and then adapts to a target domain by fine-tuning to get better performance. Dose \textit{et al.} trained a subject-independent classifier and adapted it to every single subject \cite{[49]}. Hang \textit{et al.} improved the classification performance on target domain with the deep features extracted from source domain's raw EEG signals, considering that the data of target subject was the target domain and the data of other subjects was the source domain \cite{[50]}. Zhao \textit{et al.} further added a discriminator in an end-to-end model for learning well of the shared features of the source domain and target domain \cite{[51]}.

In summary, we use the CS-GAN for EEG data augmentation with a small number of samples from the target subject, and then use these augmentation data to enhance the cross-subject classification ability through adaptive training.

\section{Methods}
\begin{figure}[h]
    \centering
    \includegraphics[width=0.9\linewidth]{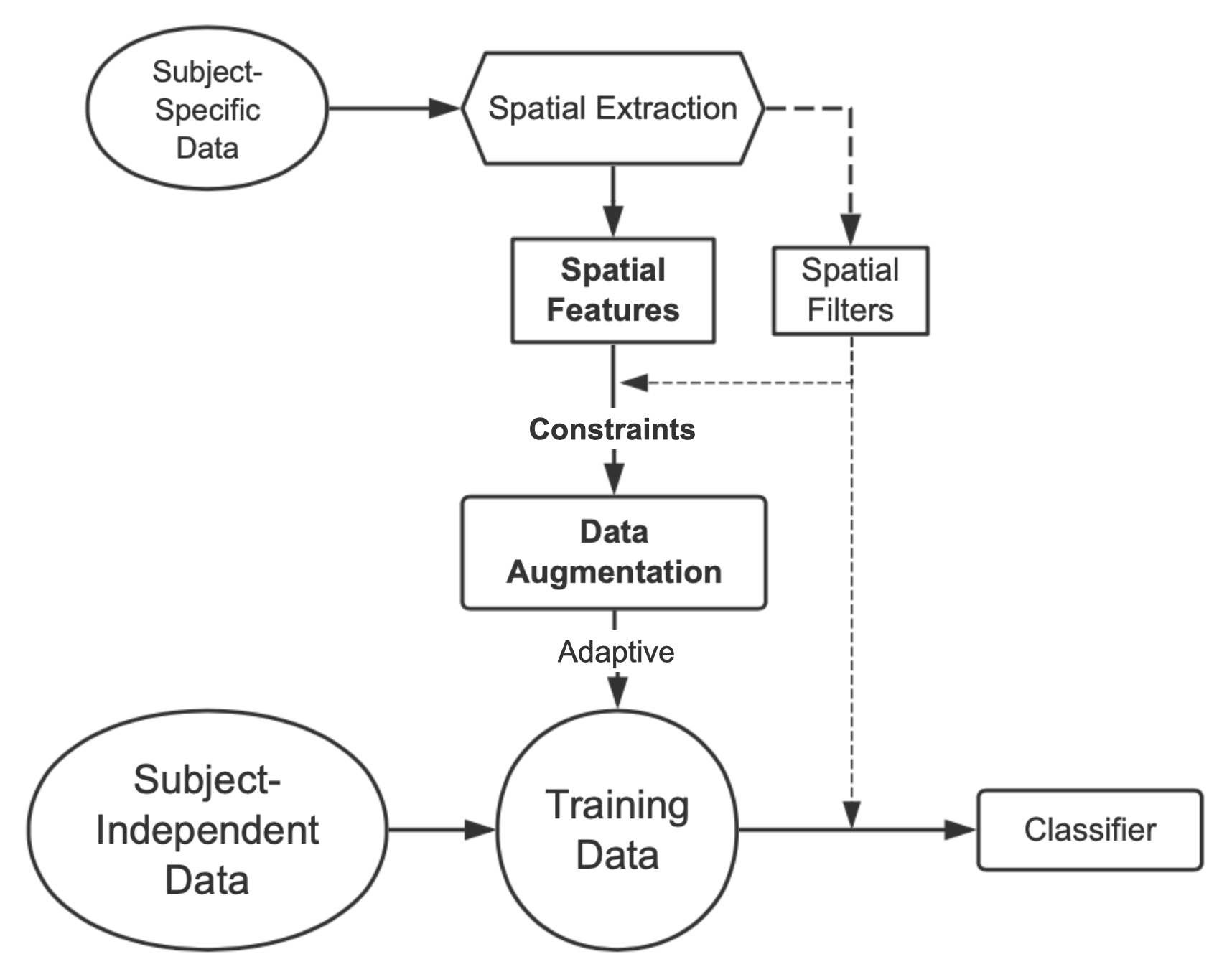}
    \caption{The overall framework of cross-subject EEG signal classification with data augmentation.}
    \label{fig:1}
\end{figure}

In the actual use of BCI, too many subject-specific data are need for calibration, under the case where good classification results cannot be accomplished with only subject-independent data. Here, we propose CS-GAN, an EEG signal augmentation method focusing on spatial enhancement, which employs a small amount of subject-specific data to generate data with the same characteristics of the original signal and improve the cross-subject performance in EEG classification tasks. The overall framework of the augmented classification with CS-GAN is shown in Fig. \ref{fig:1}. Firstly, subject-specific data is processed to obtain the spatial features and spatial filters. Then the spatial features and are used as the constraints in CS-GAN for data augmentation and enhance the discrimination between categories, and then many generated data are introduced to subject-independent data for adaptive training. The spatial filters are also used in CS-GAN, and finally applied to the training set of the classifier.  

\subsection{Data Description}
Dataset 2a of BCI competition IV, provided by Graz University of Technology is used for demonstrating our method \cite{[52]}. The dataset contains EEG data sampled at 250 Hz from 9 subjects, which were collected with 22 channels on four different motor imagery tasks, including the imagination of movement of the left hand, right hand, both feet, and tongue. Two sessions recorded on different days were obtained for each subject with 288 trails in one session, 72 trails per motor imagery task. All the signals were bandpass filtered between 0.5 Hz and 100 Hz (with the 50 Hz notch filter enabled).

\subsection{Pre-processing}
Four-second segmentation of a trail ([2, 6] \textit{second}) was taken as a sample from the cue beginning to the end. 'NaN' were replaced with 0. The EEG signals were further filtered between 4 Hz and 40 Hz to include $\mu$ band and $\beta$ band \cite{[53]}.

The 578 samples ('T' and 'E') of one subject were shuffled, of which 516 constituted training set and 50 constituted test set to evaluate the performance of the CNN we proposed. However, only 100 samples in 'T' of one subject were inputted to CS-GAN, demonstrating the good ability to generate new data. In the final cross-subject test, the 'T' data of 8 subjects were used for training, and the other 188 data of the remaining one subject was used for the test.

The z-score standardization was performed to reduce the nonstationarity and fluctuation.The standardization can be formulated as
\begin{equation}
    X'=\frac{X-\mu}{\sqrt{\sigma^2}}
\end{equation}
where $X'$ and $X$ denote the standardized and input filtered signal. $\mu$ and $\sigma^2$ represent the mean value and variance calculated with the training set, using which the signals become normally distributed with a mean of 0 and a standard deviation of 1. Then the mean and variance are applied to the test set.

\subsection{Spatial Features and Filter}
The spatial features of EEG, especially the correlation between different channels, are often ignored. Therefore, in this part, we extracted the spatial features of the EEG samples and constructed a mapping space, in which different categories of motor imagery could be better distinguished.  

Because of the multi-classification task we faced, a modified one-versus-rest (OVR) strategy was adopted to overcome the shortcoming of traditional usage of spatial filter, which only separated two categories. OVR means that the multi-classification was transformed into multiple bi-classification, consisting of one class and the remaining classes. We calculated each sample's covariance matrix for one bi-classification in four of our whole task as 
\begin{equation}
    cov(X) = \frac{XX^T}{trace(XX^T)} 
\end{equation}
where $cov()$ means to obtain the covariance matrix, and $trace()$ means to calculate the trace of a matrix. $X$ is the EEG sample, which can be expressed as $C\times T$, $C$ is the number of the channels, and $T$ is the sample length. The shape of $X$ in this task is $22\times1000$.

After that, all samples in a category were averaged to obtain two covariance matrices $R_1$ and $R_2$, $R_1$ is the 'one', and $R_2$ is the 'rest', which depicted the spatial relationship between channels of the two categories, respectively. The reason is that covariance is a measure of the joint variability of two random variables, and the covariance matrix contains the covariance of every two rows in the original sample. To utilize the spatial features derived from the covariance matrix, the Euclidean distance from each sample's covariance matrix to $R_1$ was calculated as
\begin{equation}
    \begin{aligned}
        Dis & = \left\| cov(X)-R_1 \right\| \\
        & = \sqrt{\sum_{i=1}^{C} \sum_{j=1}^{C}[cov(X)_{ij}-R_{1ij}]^2}
    \end{aligned}
\end{equation}
where $Dis$ denotes the Euclidean distance between $cov(X)\in \mathbb R^{C\times C}$ and $R_1\in \mathbb R^{C\times C}$. The mean $Dis_{mean}$ and standard deviation $Dis_{std}$ of all $Dis$, and $R_1$ were saved for later use. Next, a common space $R$ was obtained by
\begin{equation}
    R = R_1 + R_2
\end{equation}
And eigendecomposition was conducted as 
\begin{equation}
    R = U\Lambda U^T
\end{equation}
where $U$ and $\Lambda$ (sorted in descending) represent the eigenvectors and the eigenvalues, respectively. Then the whitening matrix $P$ of $R$ was obtained: 
\begin{equation}
    P = \sqrt{\Lambda^{-1}} U^T
\end{equation}
with which $R_1$ and $R_2$ were transformed as 
\begin{equation}
    \begin{aligned}
        & S_1 = P R_1 P^T \\
        & S_2 = P R_2 P_T \\
    \end{aligned}
\end{equation}
The orthogonal diagonalization of $S_2$ was obtained as
\begin{equation}
    S_2 = B \Lambda_S B^T
\end{equation}
where $B$ and $\Lambda_S$ (sorted in ascending) are the eigenvectors and eigenvalues of $S_1$. Due to orthogonality, there was 
\begin{equation}
    B^T B = I
\end{equation}
\begin{equation}
    \begin{aligned}
        I & = B^T \sqrt{\Lambda^{-1}} U^T U\Lambda U^T (\sqrt{\Lambda^{-1}} U^T)^T B \\
        & = B^T P R P^T B \\
        & = B^T P R_1 P^T B + B^T P R_2 P^T B \\
        & = B^T P R_1 P^T B + \Lambda_S \\
    \end{aligned}
\end{equation}
where $B^T P R_1 P^T B$ is a diagonal matrix, denoted as $\Lambda'_S$ with a value of $I-\Lambda_S$, which means that $B^T P$ diagonalizes both $R_1$ and $R_2$. We could see that the values of $\Lambda'_S$ become larger when the values of $\Lambda_S$ becomes smaller. $B$ and $P$ were also saved for later use in CS-GAN. Moreover, $P^T B$ was considered as a spatial filter, with which the difference between the 'one' and the 'rest' was maximized. We obtained four sub-filters here because of the four categories we had, and the first four columns of each sub-filter were token corresponding to the largest four eigenvalues of $\Lambda'_S$, to reduce the computational complexity. Then the final spatial filter $W$ was completed by stacking the four sub-filter and could be used as
\begin{equation}
    Z = W' X;
\end{equation}
where $Z$ is the processed sample. In this framework, the spatial filter $W$ with a shape of $channel\times 16$ was saved for CS-GAN. The number of columns of the spatial filter could be chosen as actual needs.

\subsection{CS-GAN}
\begin{figure*}[h]
\centering
\includegraphics[width=\linewidth]{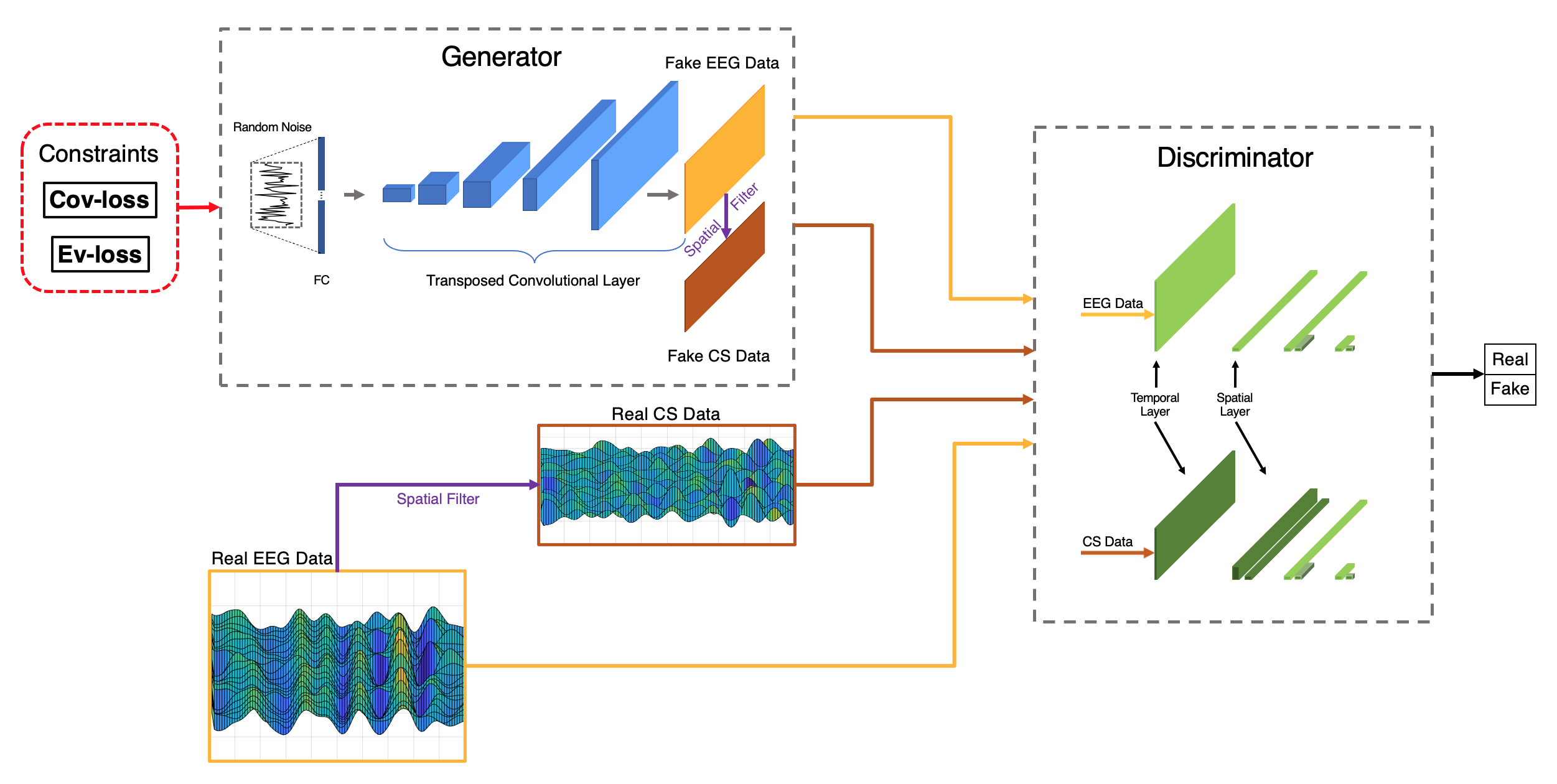}
\caption{The architecture of CS-GAN, which contains a generator and a discriminator. The discriminator has EEG module and CS-module. The cov-loss and ev-loss are used as constraints. CS-module, cov-loss and ev-loss are used together to maintain EEG signal's spatial features and enhance the spatial difference between different categories.}
\label{fig:2}
\end{figure*}
The most important part of this article is to propose a novel EEG data augmentation method based on GANs, which effectively maintain the characteristics of the original signal while enhancing the spatial features and discrimination between different categories. 

GAN is a training strategy that constructs a competition between two networks named generator $G$ and discriminator $D$. To illustrate with an analogy, it is a game between counterfeiter and police, where the counterfeiter tries to deceive the police with fake paintings. In actual use, artificial data generated by $G$ from random noise are inputted to $D$ along with real data, and $D$ identifies whether it is real or fake. $G$ and $D$ finally reach equilibrium after enough adversarial training. At this point, the generated data captures the distribution of real data, and cannot be distinguished by the discriminator.  

The architecture of CS-GAN is given in Fig. \ref{fig:2}. As we can see, it also includes a generator part and a discriminator part. But different from other research, we have innovatively expanded the discriminator into two modules to distinguish EEG data and common spatial data (CS data), respectively. Besides, $R_1$, $Dis_{mean}$, $Dis_{std}$, $P$ and $B$ obtained in the previous step were used to constrain the generator and better retain EEG samples' spatial features. The detailed design of CS-GAN is given as follows. 

\subsubsection{Generator}
The generator was employed to generate fake EEG signals similar to the real EEG signals of different motor imagery. We inputted a piece of normally distributed random noise ($1\times 1600$) to the generator and introduced one fully-connected layer and five transposed convolutional layers to up-sample the input to the same size as the original sample ($channel\times sample\ point$). Note that the input size was longer than the input of common GANs, due to the scale of EEG samples was also more extensive. 

The generator's network structure is shown in Table \ref{tab:1}, where $FC$ is the fully-connected layer, and $ConvTrans$ is the transposed convolutional layer. It is worth noting that our method used 2-dimensional convolution to learn the deep relationship among channels, some of the spatial features of EEG. Batch normalization was applied behind the first four transposed convolutional layers. The activation function of the fully-connected layers and the transposed convolutional layers except the last layer was LeakyReLU with a negative slope of 0.2.
\begin{table}[ht]
\scriptsize
\begin{center}
	\caption{The network structure of the generator.}
	\label{table}
	\setlength{\tabcolsep}{3pt} 
	\renewcommand\arraystretch{1.5} 
	\begin{tabular}{|c|p{1cm}<{\centering}|p{1cm}<{\centering}|p{1cm}<{\centering}|p{1cm}<{\centering}|}
		\hline
		Layers & Input & Output & Kernel & stride \\ \hline
		\multirow{1}*{$FC$} 
		& 1600 & 256000 & - & -  \\ \hline
		\multirow{5}*{$ConvTrans$} & 128 & 128 & (3, 15) & (1, 3) \\ \cline{2-5}  
		& 128 & 128 & (3, 13) & (1, 3) \\ \cline{2-5}
		& 128 & 64 & (3, 5) & (1, 2) \\ \cline{2-5}
		& 64 & 32 & (4, 5) & (2, 1) \\ \cline{2-5}
		& 32 & 1 & (1, 2) & (1, 1) \\ \hline
	\end{tabular}
	\label{tab:1}
\end{center}
\end{table}

\subsubsection{Discriminator}
Compared with the past usage, we have considerably changed the discriminator part, the purpose of which is to make the generated data retain the spatial features of original data, through more targeted adversarial training. The usual practice is to input original real data and generated fake data into the discriminator. In our method, the spatial filters obtained from the previous step were introduced firstly to process the real EEG data and fake EEG data. Then the results (called real CS data and fake CS data) were input to the discriminator together with real EEG data and fake EEG data. Two modules distinguished whether the EEG data and CS data were real or fake. 

The discriminator's network structure is given in Table \ref{tab:2}, where $Conv$ is the convolutional layer, $Maxpool$ is the max-pooling layer, and $FC$ is the fully-connected layer. $Temporal\ Conv$ and $Spatial\ Conv$ were used to extract the temporal features and spatial features separately, with which the distinguishing ability of the discriminator was enhanced. The only difference between EEG module and CS-module was that the kernel size of $Spatial\ Conv$ in EEG module was $(channels, 1)$. And there was an extra layer to separate the channels according to four sub-filter for four categories in $Spatial\ Conv$ of the CS-module. LeakyReLU with a negative slope of 0.2 was applied as the activation function after each convolutional layer, but no batch normalization. Finally, two fully-connected layers were used to obtain the probability of identifying EEG data and CS data, respectively. In this way, not only the EEG signals conformed to the original distribution would be generated, but the good feature characteristics and separability brought by the spatial filter was maintained. 
\begin{table}[ht]
\scriptsize
\begin{center}
	\caption{The network structure of the discriminator.}
	\label{table}
	\setlength{\tabcolsep}{3pt} 
	\renewcommand\arraystretch{1.5} 
	\begin{tabular}{|c|c|p{1cm}<{\centering}|p{1cm}<{\centering}|p{1cm}<{\centering}|p{1cm}<{\centering}|}
		\hline 
		Modules & Layers & Input & Output & Kernel & stride \\ \hline
		\multirow{7}*{EEG} & $Temporal\ Conv$ & 1 & 10 & (1, 23) & (1, 1) \\ \cline{2-6}
		& $Spatial\ Conv$ & 10 & 30 & (22, 1) & (1, 1) \\ \cline{2-6}
		& $Conv$ & 30 & 30 & (1, 17) & (1, 1) \\ \cline{2-6}
		& $Maxpool$ & - & - & (1, 6) & (1, 6) \\ \cline{2-6}
		& $Conv$ & 30 & 30 & (1, 7) & (1, 1) \\ \cline{2-6}
		& $Maxpool$ & - & - & (1, 6) & (1, 6) \\ \cline{2-6}
		& $FC$ & 750 & 1 & - & - \\ \hline
		\multirow{8}*{CS} & $Temporal\ Conv$ & 1 & 10 & (1, 23) & (1, 1) \\ \cline{2-6}
		& \multirow{2}*{$Spatial\ Conv$} & 10 & 30 & (4, 1) & (4, 1) \\ \cline{3-6}
		& & 30 & 30 & (4, 1) & (1, 1) \\ \cline{2-6}
		& $Conv$ & 30 & 30 & (1, 17) & (1, 1) \\ \cline{2-6}
		& $Maxpool$ & - & - & (1, 6) & (1, 6) \\ \cline{2-6}
		& $Conv$ & 30 & 30 & (1, 7) & (1, 1) \\ \cline{2-6}
		& $Maxpool$ & - & - & (1, 6) & (1, 6) \\ \cline{2-6}
		& $FC$ & $750$ & 1 & - & - \\ \hline
	\end{tabular}
	\label{tab:2}
\end{center}
\end{table}

\subsubsection{Loss Function}
The loss function of CS-GAN mainly consisted of adversarial loss, covariance loss (cov-loss) and eigenvalue loss (ev-loss). The adversarial was constructed based on Wasserstein GAN with a gradient penalty to stable the training processing \cite{[30]} as
\begin{equation}
    \mathcal{L}_{adv} = E_{G(z)\sim \mathbb{P}_g}[D(G(z))]-E_{x\sim \mathbb{P}_r}[D(x)] +\lambda_{gp}GP(\hat{x})
\end{equation}
\begin{equation}
    GP(\hat{x}) = E_{\hat{x}\sim \mathbb{P}_{\hat{x}}}[(\left\| \nabla_{\hat{x}}D(\hat{x}) \right\|_2-1)^2]
\end{equation}
where $E$ is the expectation operator, $G(z)$ denotes the generated sample produced by the generator $G$ from a random noise input $z$ and $D(x)$ is the probability of sample $x$ belonging to the real. $\mathbb{P}_g$ is the generated data distribution and $\mathbb{P}_r$ is the real data distribution. $GP()$ is the gradient penalty, and $\hat{x}$ represents the data obtained by linear sampling between the generated data $G(z)$ and the real data x, where $\hat{x}=\alpha x+(1-\alpha)G(z)$ and $\alpha$ is a random value on the interval $(0,1)$.

The cov-loss was proposed to make the covariance matrix of each generated sample approximate to the real samples of its category because the spatial relationship between channels of a sample was reflected by the covariance matrix. It was calculated as
\begin{equation}
    \mathcal{L}_{cov} = \frac{abs(\left\| cov(G(z))-R_1 \right\|-Dis_{mean})}{Dis_{std}} 
\end{equation}
where $loss$ is set to 0 when $loss\leq 0$, $abs()$ means absolute value operator and $cov(G(z))$ is the covariance matrix of the generated sample. The Euclidean distance from which to the average covariance matrix of the corresponding category is calculated and made as close as possible to the original samples' distance distribution.

According to the previous steps, the eigenmatrix $\Lambda_S$ of the covariance matrix could be diagonalized by $B^T$, and the larger values of $\Lambda'_S$ corresponded to the smaller value of $\Lambda_S$. The first four columns of the spatial filter were chosen corresponding to the four largest values of $\Lambda'_S$, when obtaining the spatial filer. Therefore, we designed the eigenvalue loss as
\begin{equation}
    ev = diag(B^TP\ cov(G(z))\ P^TB)[1:4]
\end{equation}
\begin{equation}
    \mathcal{L}_{ev} = abs(log(E(ev)) 
\end{equation}
where $diag()$ is used to obtain diagonal elements, $ev$ is the largest four eigenvalues of the generated samples after transformation, and $log()$ is the natural logarithm, which is used to make the eigenvalues larger within the range of zero to one, because the diagonal elements of both $B^TPR_1P^TB$ and $B^TPR_2P^TB$ are greater than zero. Thus, the category difference of the generated samples increases with eigenvalue loss, further promoting later classification. 

The adversarial loss was improved for the discriminator part so that it could identify original EEG data and CS data simultaneously. The overall loss of the generator and the discriminator could be formulated in the following equations:
\begin{equation}
    \begin{aligned}
        \mathcal{L}_G = & -E(D(G(z))) - \lambda_{cs}E(D(WG(z))) \\
        & + \lambda_{cov}\mathcal{L}_{cov} + \lambda_{ev}\mathcal{L}_{ev}
    \end{aligned}
\end{equation}
\begin{equation}
    \begin{aligned}
        \mathcal{L}_D = & E(D(G(z)) - E(D(x)) + \lambda_{gp}GP(\hat{x}) \\
        & + E(D(WG(z)) - E(D(Wx)) + \lambda_{gp}GP(W\hat{x})
    \end{aligned}
\end{equation}
where $W$ is the spatial filter obtained for the previous step. $\lambda_{gp}$, $\lambda_{cs}$, $\lambda_{cov}$, $\lambda_{ev}$ denote the weight of gradient penalty, the loss of common spatial part, covariance loss and eigenvalue loss, separately, the values of which were determined to be $10$, $0.1$, $3$ and $10$ by comparative experiments.  
\subsection{Classifier}
We presented a general method for EEG classification, which was also used to test CS-GAN's performance. The first step was to get a spatial filter using matrix transformation and revised OVR strategy as before. The category difference was increased by projecting the original data of different categories into a new common space. The processed data had 16 channels, and every four channels was a kind of representation that was easier to be distinguished into one category. 

A CNN was designed to learn such mixed enhanced representation as TABLE \ref{tab:3}. After initializing the network parameters, a convolutional layer was used to separate the sample into four parts, with which the four categories were more manageable to be token out, respectively. Then several convolutional layers with max-pooling layers were used to extract temporal features. And there were three fully-connected layers to obtain the deep representations. The activation function was LeakyReLU with a negative slope of 0.2. Dropout with a rate of 0.3 in fully-connected layers was used to avoid overfitting. Finally, a fully-connected layer with four units and softmax function was implemented to obtain the classification probabilities of the four categories. 

\begin{table}[ht]
\scriptsize
\begin{center}
	\caption{The network structure of the CNN classifier.}
	\label{table}
	\setlength{\tabcolsep}{3pt} 
	\renewcommand\arraystretch{1.5} 
	\begin{tabular}{|c|p{1cm}<{\centering}|p{1cm}<{\centering}|p{1cm}<{\centering}|p{1cm}<{\centering}|}
		\hline 
		Layers & Input & Output & Kernel & stride \\[0.5pt] \hline
		$CS\ Conv$ & 1 & 16 & (4, 1) & (4, 1) \\ \hline
		$Conv$ & 16 & 32 & (1, 23) & (1, 3) \\ \hline
		$Conv$ & 32 & 64 & (1, 17) & (1, 1) \\ \hline
		$Maxpool$ & - & - & (1, 6) & (1, 6) \\ \hline
		$Conv$ & 64 & 128 & (1, 7) & (1, 1) \\ \hline
		$Maxpool$ & - & - & (1, 2) & (1, 2) \\ \hline
		\multirow{4}*{$FC$} & 11264 & 2048 & - & - \\ \cline{2-5}
		& 2048 & 512 & - & - \\ \cline{2-5}
		& 512 & 128 & - & - \\ \cline{2-5}
		& 128 & 4 & - & - \\ \hline
	\end{tabular}
	\label{tab:3}
\end{center}
\end{table}

\section{Experiments and Results}
An approach has been proposed to improve the cross-subject practicability of motor imagery EEG classification along with the idea of adaptive training. A small amount of subject-specific data is augmented with CS-GAN to introduce individual specific information into the subject-independent data.  

In this section, we are going to validate the ability of this method to improve the cross-subject classification from multiple perspectives. Firstly, we confirm the effectiveness of this method by comparing whether to add augmented data. Secondly, we compare the improvement of our method and other great augmentation methods in cross-subject classification and the ablation test of CS-GAN. Additionally, our benchmark for classification is compared with several great methods to ensure the rationality of the previous verification. The generated EEG signals are also compared with original signals in the temporal domain and frequency domain. Finally, we test the generalization ability of our method on dataset 2b of BCI competition IV. 

\subsection{Experiment Settings}
Our framework was implemented with the PyTorch library in Python 3.6 on a server with Intel Xeon CPU and Geforce 2080Ti GPU. For the dataset, the three electrooculography (EOG) channels were directly discarded without artifact removing operation. In CS-GAN, Adam was used as the optimizer with a learning rate of 0.0001, $\beta_1$ of 0.1 and $\beta_2$ of 0.999. The network parameters were updated after every batch with a size of 5. \textit{kaiming} uniform initialization was employed. In the classifier, $\beta_1$ changed to 0.9 and the batch size to 50. 

The 'T' session of nine subjects in BCI competition IV dataset 2a was chosen for the experiments so that each subject had 288 samples, 72 samples for each category. Besides, 100 data was randomly chosen from the 288 samples to test the extreme performance with a few data. We know that GANs usually need a lot of data for training, but only 16 to 31 data were used for augmenting one category with CS-GAN. Moreover, the paired t-test was employed for statistical analysis.

\subsection{Data Augmentation for Cross-Subject Classification}
Leave-one-subject-out (LOO) validation was firstly performed, and the results are shown in Table \ref{tab:4}. As we can see, due to the inherent non-stationary of EEG signal, if only the data of eight subjects is used for training and the data of the other one subject is used for the test, very poor average classification accuracy of 52.12\% with a standard deviation of 14.08\% is obtained which cannot be used in practice. 

\begin{table*}[h]
\scriptsize
\begin{center}
	\caption{classification accuracy (in percentage \%) under different augmentation situations}
	\label{table}
	\setlength{\tabcolsep}{3pt} 
	\renewcommand\arraystretch{1.5} 
	\begin{tabular}{|c|c|c|c|c|c|c|c|c|c|c|}
		\hline 
		\ & S01 & S02 & S03 & S04 & S05 & S06 & S07 & S08 & S09 & Accuracy
		\\ \hline
		Leave-one-subject-out & 69.10 & 36.81 & 60.76 & 45.83 & 33.33 & 42.71 & 45.83 & 65.97 & 68.75 & 52.12$\pm$14.08 \\ \hline
		adapt 100 real samples & 72.87 & 42.55 & 70.74 & 53.72 & 40.96 & 41.49 & 66.49 & 76.06 & 69.68 & 59.40$\pm$14.67 \\ \hline
		adapt 100 fake samples & 77.13 & 38.30 & 69.15 & 51.60 & 39.36 & 39.89 & 65.96 & 78.72 & 62.77 & 58.10$\pm$16.24 \\ \hline
		adapt 3000 fake samples & 81.91 & 53.19 & 79.26 & 60.11 & 44.68 & 49.47 & 80.32 & 84.04 & 78.72 & \textbf{67.97$\pm$15.86} \\
		\hline
	\end{tabular}
	\label{tab:4}
\end{center}
\end{table*}

As far as we know, there is no widely accepted method to directly use any method to achieve good enough LOO classification results, because we have no reasonable basis for distinguishing the motor imagery related information and identity related information. We cannot introduce the information reflects the data distribution of a specific subject as well. Therefore, some research tries to pre-train with the subject-independent data and adaptively fine-tune with the subject-specific data. As Table \ref{tab:3} presents, the classification accuracy is 59.40\% when adapting 100 real subject-specific data into the subject-independent data, which is significantly higher than that of LOO (\textit{p-value}\ \textless\ 0.05). With the 100 data, the classifier could better perceive the characteristics of the specific subject and separate the four categories of motor imagery.  We also have a test with 100 fake data generated by CS-GAN instead of real data. The result shows that the average classification accuracy of adapting 100 fake data is only 1.30\% (\textit{p-value}\ \textgreater\ 0.05) lower than adapting 100 real data.  

Naturally, we try to introduce more fake data generated by CS-GAN to provide more subject-specific information to the classifier. 750 each category, a total of 3000 augmented samples are adapt into earlier, which are just more than the subject-independent data. It can be seen that sufficient data greatly improve classification performance. The average result increase by 8.57\% (\textit{p-value}\ \textless\ 0.001) compared to adapting 100 real data and 15.85\% (\textit{p-value}\ \textless\ 0.001) compare to LOO. So far, it has confirmed the method adding augmentation data for adaptive training is effective for the cross-subject situation with a significant improvement of the classification accuracy for specific subject. 

\begin{figure}[ht]
\centering
\includegraphics[width=\linewidth]{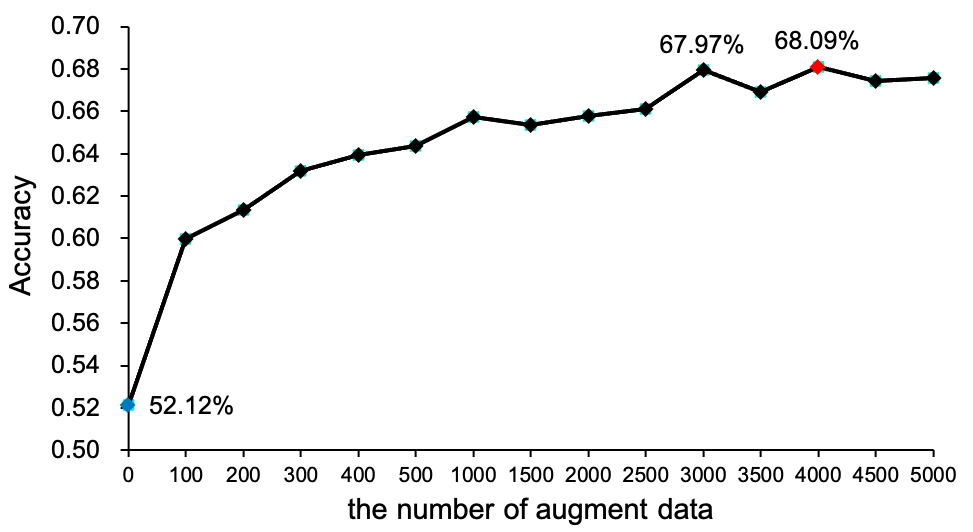}
\caption{The average EEG classification accuracy with the different number of augmentation data.}
\label{fig:3}
\end{figure}

More results are obtained by adapting the different number of fake data shown in Fig. \ref{fig:3}. In the graph, the average classification accuracy of the nine subjects is dramatically improved when we continue to introduce more fake data at the beginning. After adding more than 1000 fake data, the accuracy gradually stabilizes and reaches a maximum value of 68.09$\pm$15.53\% at 4000, which is 15.97\% higher than the result without augmentation data (\textit{p-value}\ \textless\ 0.001). It can also be seen that the results begin to fluctuate and there is no visible upward trend after 3000.

\subsection{Some Augmentation Methods and Ablation Study}
In this part, we compare CS-GAN with some other great augmentation methods and perform ablation study to further ensure the CS-GAN's superiority and the effectiveness of the additional CS-module, cov-loss and ev-loss. The input was the same as CS-GAN, and 3000 fake data generated by different ways were introduced for estimation, respectively. The average accuracy was obtained from the results of nine subjects. 

\subsubsection{Adding Gaussian Noise}
Adding Gaussian noise to the pre-processed data with normal distribution is a common method which will not change the original distribution of the EEG signals. The standard deviation of the noise was chosen to be 0.2, according to the optimal result obtained in \cite{[18]}. 

\subsubsection{Segmentation and Recombination (S\&R)}
We implemented S\&R, a widely accepted method, to generate augmentation data. EEG trials of the same category were divided into several segments, which would be concatenated randomly maintaining the order in the trial. A trial was segmented into eight segments, as suggested in \cite{[54]}. 

\subsubsection{Variational Auto-Encoder (VAE)}
Generative models that appear in recent years have also been compared. The first was VAE inspired by \cite{[39]}, which employed an encoder consisted of a 1-dimensional convolutional layer with max-pooling layer and two fully-connected layers, and a decoder consisted of a fully-connected layer and three transposed convolutional layers. The encoder projected the EEG data to latent vectors, using which the decoder reconstructed the input. 

\subsubsection{Deep Convolutional GAN (DCGAN)}
We also implemented DCGANs which showed remarkable performance in \cite{[40]}, one of the few works to apply GANs for EEG data augmentation. The generator used two fully-connected layers and up-sampling followed by convolutional layers. The discriminator consisted of pairs of convolutional and max-pooling layers, followed by a fully-connected layer. Adam with a learning rate of 0.0001 and $\beta_1$ of 0.2 was employed for optimization.

\subsubsection{Ablating All}
The difference between our CS-GAN and other GANs lies in the CS-module, cov-loss and ev-loss, in addition to the well-designed network structure with particular parts for temporal and spatial information. Therefore, the ablation tests were conducted from the above perspectives. Firstly, the CS-module was completely removed, and $\lambda_{cov}$ as well as $\lambda_{ev}$ was set to be $0$. It can be said to be a kind of DCGAN with Wasserstein loss.

\subsubsection{Ablating Common Spatial Module (CS-M), covariance loss (cov-loss), eigenvalue loss (ev-loss)}
The role of CS-module was to maintain EEG data's spatial characteristics for classification by controlling the projected data, since we had constructed a new subspace where the four categories had greater discrimination. Then the cov-loss was employed to further holding the spatial pattern of EEG signals, and the ev-loss was used to improve the discrimination between different categories. Therefore, we tested the classification performance using the fake data generated with no CS-module, no cov-loss and no ev-loss separately. 

\begin{figure}[ht]
\centering
\includegraphics[width=\linewidth]{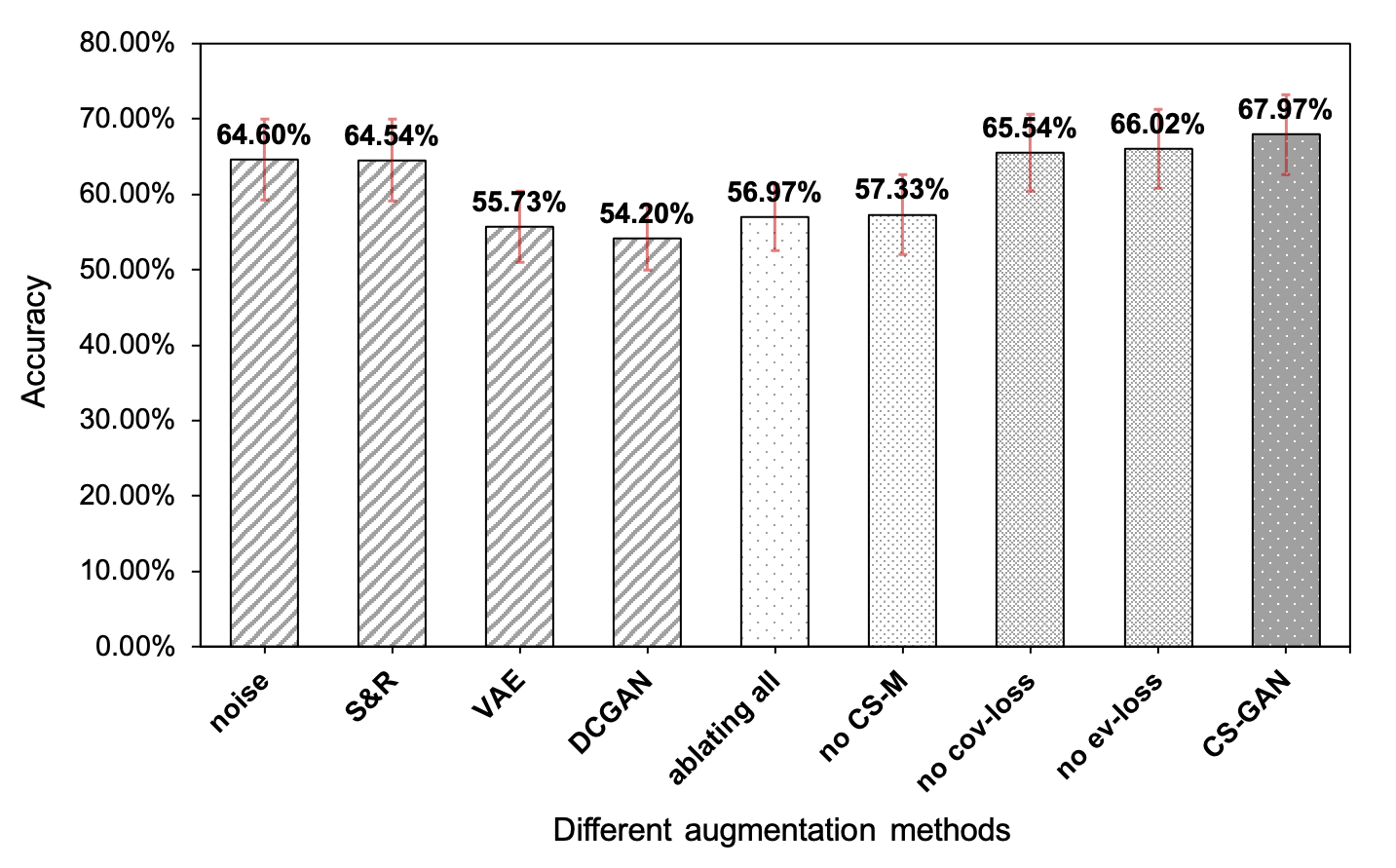}
\caption{The average classification accuracy using different augmentation methods, including adding Gaussian noise, segmentation and reconstruction, variational auto-encoder, deep convolutional GAN and ablation test.}
\label{fig:4}
\end{figure}

These competitive methods have been used to generated fake data for augmentation, and the classification results are shown in Fig. \ref{fig:4}. As we can see, these two commonly used data augmentation methods, adding Gaussian noise and S\&R, indeed have very good results. However, the introduced noise reduces the signal-to-noise ratio of the EEG, which itself has a lot of noise. Segmenting the samples and randomly reconstructing destroys each sample's inherent consistency because the latent period is usually difficult to keep the same. Therefore, such shortcomings make the results of these two methods unsatisfactory and lower than the result of CS-GAN (Noise: 3.37\%, \textit{p-value}\ \textless\ 0.001; S\&R: 3.43\%, \textit{p-value}\ \textless\ 0.01).    

Then there are two popular generative models, VAE and DCGAN, which some researchers have also used to generate EEG signals. However, the results are obviously lower than the results of CS-GAN (VAE: 12.24\%, \textit{p-value}\ \textless\ 0.001; DCGAN: 13.77\%, \textit{p-value}\ \textless\ 0.001). That is not to say these methods are not good. In fact, they have achieved wonderful results. However, in the cross-subject situation, the data can be used to train the generative data is too small, compared to the amount required by other methods (50 for VAE \cite{[39]} and 520 for DCGAN \cite{[40]}).

Actually, the test that ablates the CS-module, cov-loss and ev-loss, is also a form of DCGAN. It has been shown that even if the network structure deals the spatial and temporal information, the improvement is still not enough (ablating all: 11.00\%, \textit{p-value}\ \textless\ 0.001). Next are the results of other ablation tests. We can see that when there is no CS-module, the quality of the generated signals is greatly reduced (no CS-M: 10.64\%, \textit{p-value}\ \textless\ 0.001). The additional CS-module drives generated samples to preserve the spatial features and category discrimination because the conversion of common space is realized with the spatial features and category difference enhancement. 

Furthermore, the cov-loss, a constraint to further maintain the spatial similarity of the generated data and original data, is removed from the complete CS-GAN. Obviously, the reduced performance illustrates that it has improved the generated data quality (no cov-loss: 2.43\%, \textit{p-value}\ \textless\ 0.01), making it more conducive to augmentation. The same test has also applied to ev-loss, which makes different categories more distinguishable, after ensuring high-quality generation with other parts. The performance degradation is also significant, refer to the result without ev-loss (no ev-loss: 1.95\%, \textit{p-value}\ \textless\ 0.01).

Through the above results, it can be confirmed that our method, CS-GAN, has sufficient advantages compared to other data augmentation methods. In particular, better results has been achieved with a tiny amount of training data. This method has remarkable meaning for EEG classification and other problem with a similar condition. Moreover, ablation study helps us verify the effectiveness of the  three highlighted parts, CS-module, cov-loss and ev-loss. These method innovations have indeed shown the desired effect as mathematical reasoning. It is worth mentioning that these tests also prove the significance of focusing on the spatial features of EEG signals, which has long-term data like speech but more complicated relationship between channels and is difficult to be handled with common neural networks directly.  

\subsection{Quality of The Generated Data}
After confirming the improvement of cross-subject classification performance with CS-GAN, we tried to evaluate the quality of the generated signals from the perspective of time domain, frequency domain and spatial domain. Since the CS-GAN model for each category of each subject is parallel, subject 1's training data and generated data of right-hand movement were averaged separately for visualization. 

\begin{figure}[ht]
\centering
\includegraphics[width=\linewidth]{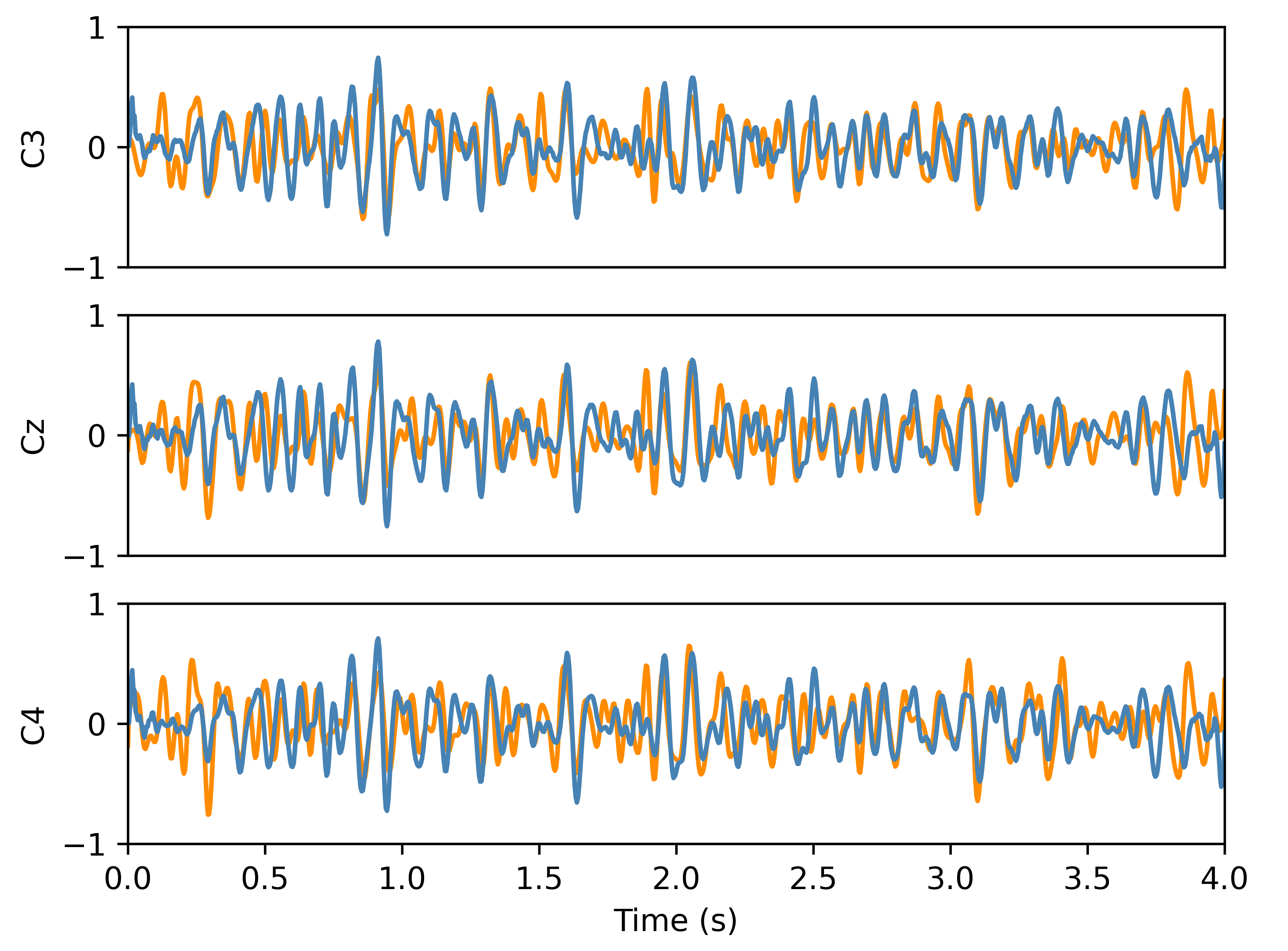}
\caption{Comparison channel C3, Cz and C4 of original real data and generated fake data. The real data is represented by orange, and the fake data is represented by blue.}
\label{fig:5}
\end{figure}

Firstly, the three main channels on the motor area, C3, Cz and C4 were chosen as Fig. \ref{fig:5} to compare the original real data and generated fake data in the time domain \cite{[55]}. We put the real signal in orange and the fake signal in blue on the same axis. It could be seen that the temporal distribution of the generated data is similar to the original data. Both mean and range are relatively close. There are some mismatches at the beginning and the end, due to the large kernel of the first two convolutional layers in the generator. Moreover, our spatial enhancement of the relationship between channels also makes the three channels more relevant, which is an attribute consistent with the real signal.

\begin{figure}[ht]
\centering
\includegraphics[width=\linewidth]{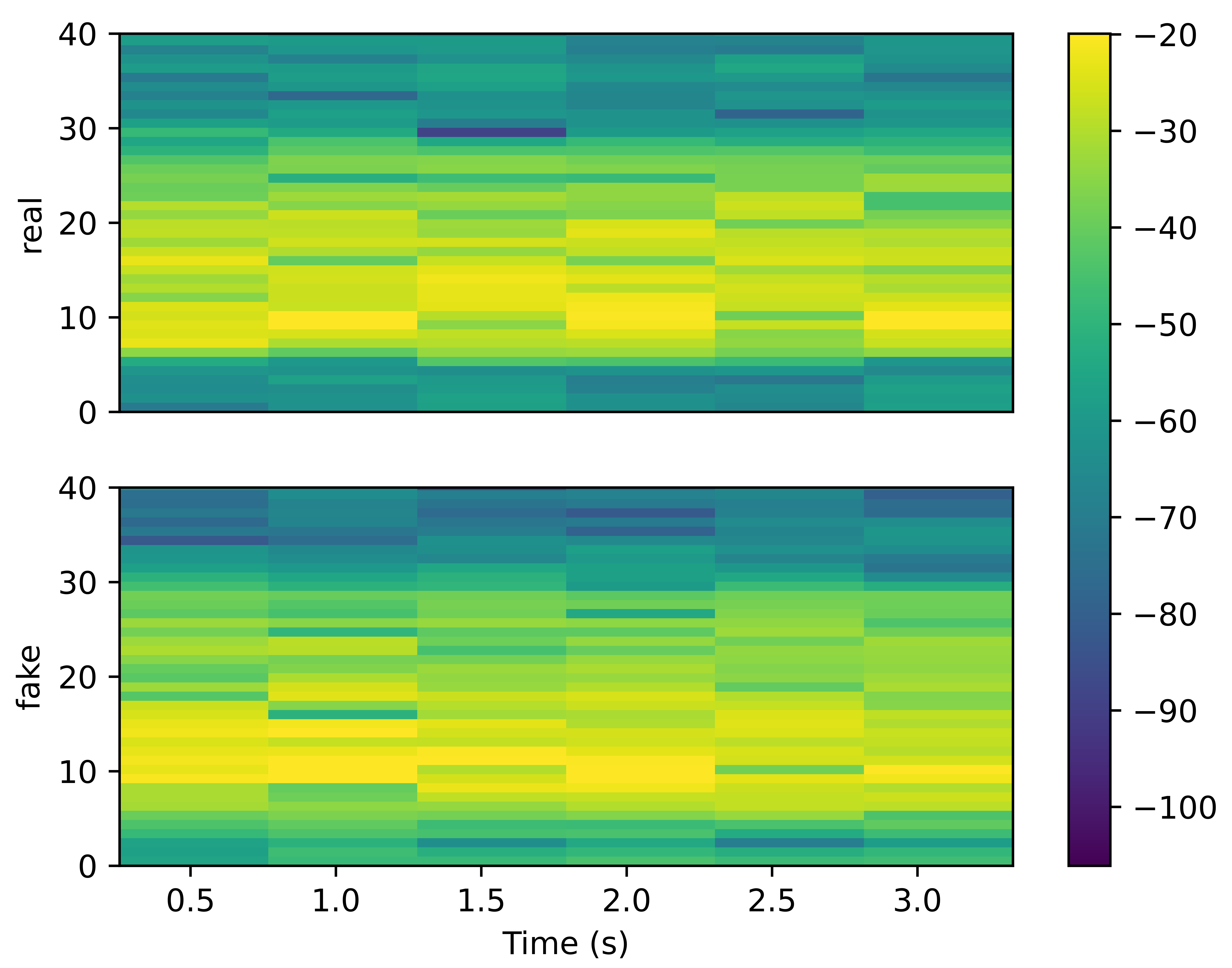}
\caption{Comparison the spectrogram of the original real data and generated fake data, of which the 22 channels are averaged. The above is real, and the below is fake. The unit of the vertical axis is Hz, and of the colorbar is dB.}
\label{fig:6}
\end{figure}

Secondly, we averaged the 22 channels of the samples and drew the spectrogram of the real and fake data in Fig. \ref{fig:6} to show the power spectral density, with which we can compare the original and generated signals by measuring the power content versus frequency. The spectrogram display 4-40 Hz as the preprocessing. It can be seen that generated signal shows higher power at the frequency of where the original signal power is higher. Especially in the range of 5 to 25 Hz, the power distribution over the entire time is close.

\begin{figure}[ht]
\centering
\includegraphics[width=\linewidth]{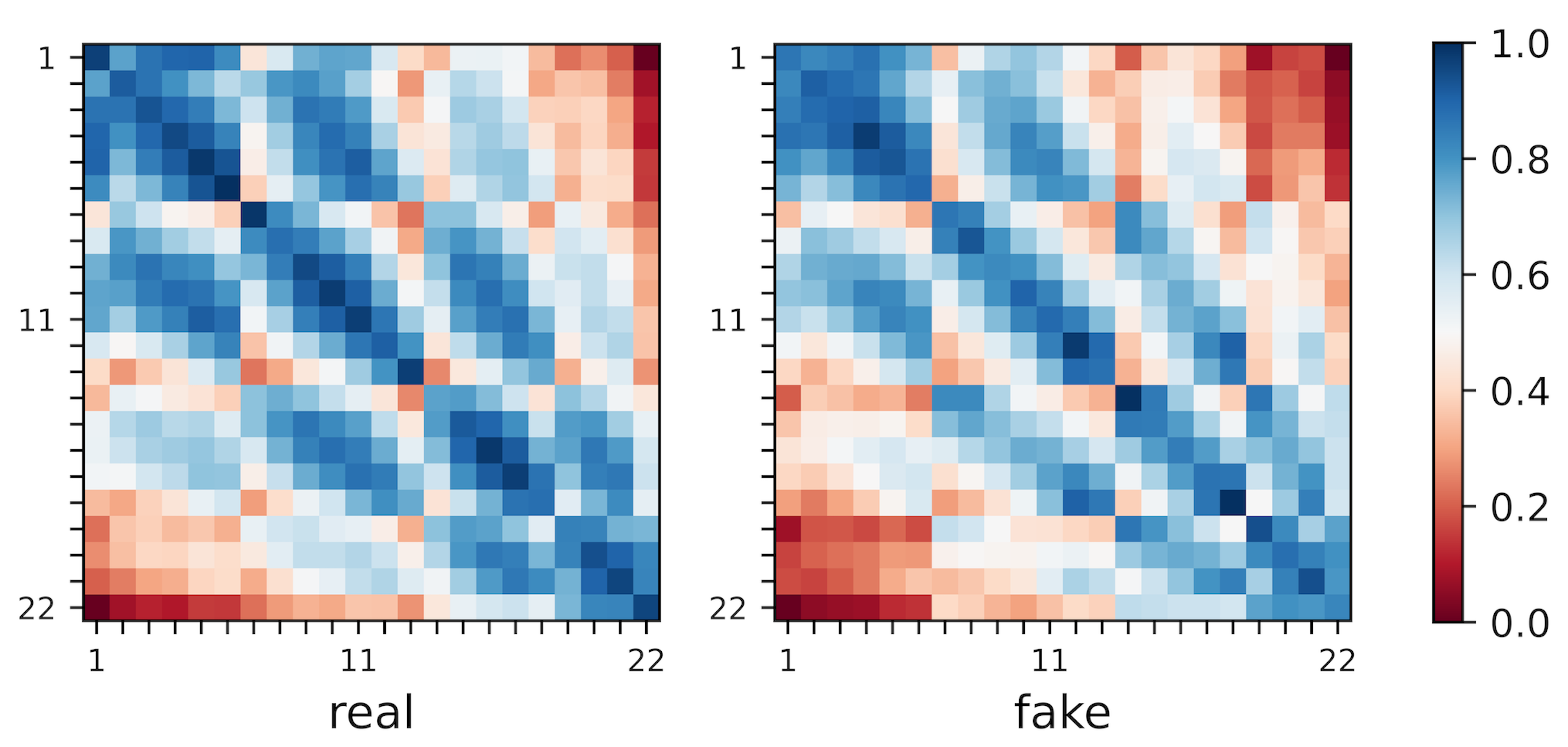}
\caption{Comparison of the heat map of the covariance matrix of original real data and generated fake data to show the relationship between channels. The coordinates from top to bottom and left to right refer to channel 1 to channel 22. Each small square represents the covariance between two channels.}
\label{fig:7}
\end{figure}

Thirdly, we focused on the spatial distribution of the generated data and used heat maps to plot the normalized covariance matrix of the original real data and generated fake data in Fig. \ref{fig:7}. This way is used to observe the relationship between the channels since the covariance matrix reflects the relationship between the rows of the data. From the heat maps, we can see that the relationship between adjacent EEG channels is well maintained, which shows that the spatial distribution of the generated data is consistent with the original data. So far, we can be confirmed that we have indeed generated fake data of sufficient quality for our augmentation.

\subsection{Classification Benchmark}
To verify the performance of the CS-GAN for cross-subject BCI classification, we also designed a multi-classification method as a benchmark. The EEG samples were transformed into a new subspace with more category discrimination and then inputted to a CNN. Here we compare with other good methods to make sure our classification method is available. It should be noted that most of the research train one classifier for each subject's data in Dataset 2a of BCI competition IV. So we do the same thing, taking about one-tenth (50) of one subject's data as the test data and the rest as the training data. The comparison results are given in Table \ref{tab:5}.

\begin{table*}[h]
\scriptsize
\begin{center}
	\caption{single-subject classification accuracy with different methods on dataset 2a of BCI competition IV}
	\label{table}
	\setlength{\tabcolsep}{3pt} 
	\renewcommand\arraystretch{1.5} 
	\begin{tabular}{|c|c|c|c|c|c|c|c|c|c|c|}
		\hline 
		\ & S01 & S02 & S03 & S04 & S05 & S06 & S07 & S08 & S09 & Accuracy (kappa)
		\\ \hline
		Multi-Branch 3D \cite{[41]} & 77.40 & 60.14 & 82.93 & 72.29 & 75.84 & 68.99 & 76.04 & 76.85 & 84.66 & 75.02 (0.6669) \\ \hline
		TSSM+LDA \cite{[56]} & 81.80 & 62.50 & 88.80 & 63.70 & 62.90 & 58.50 & 86.60 & 85.10 & 90.00 & 75.50 (0.6733) \\ \hline
		Envelop+CNN \cite{[57]} & 85.23 & 69.73 & 90.15 & 65.57 & 77.42 & 52.41 & 93.68 & 90.04 & 84.75 & 78.78 (0.7171)\\ \hline
		Functional Brain Network \cite{[58]} & 82.80 & 65.50 & 87.90 & 77.60 & 72.40 & \textbf{70.70} & 82.80 & 87.90 & 89.70 & 79.70 (0.7293)\\ \hline
		Discriminative Feature Learning \cite{[59]} & \textbf{91.31} & 71.62 & \textbf{92.32} & 78.38 & \textbf{80.10} & 61.62 & 92.63 & \textbf{90.30} & 78.38 & 81.85 (0.7580)\\ \hline
		\textbf{Ours} & 90.00 & \textbf{75.00} & 90.00 & \textbf{85.00} & 58.33 & 63.33 & \textbf{95.00} & 90.00 & \textbf{100.00} & \textbf{82.96 (0.7728)} \\
		\hline
	\end{tabular}
	\label{tab:5}
\end{center}
\end{table*}

For evaluation, we employ the accuracy and kappa value, which measures the accuracy occurring by chance. From the table, it could be seen that our method is competitive in average results compared to other state-of-the-art methods. The result of each subject also has superiority, except subject 5. Further, we can be sure that there is no problem using our classification methods as a benchmark to verify the effect of data augmentation.  

Obviously, this condition has enough subject-specific data for training (more than five times data as much as we used to train CS-GAN). As the previous experiments, the classification result obtained without subject-specific data is 30.84\% lower than the result with enough subject-specific data, and the result obtained with fewer data but no augmentation is 23.56\% lower. 

\subsection{Generalization ability of the Proposed Method}
Similar tests were also conducted on dataset 2b of BCI competition IV, which has two categories of 3-channel motor imagery data. The first session with 120 trials of each subject was chosen and separated to a training set of 50 trials and a test set of 70 trials. Twenty-five samples were used to train the CS-GAN and obtain the augmentation data of one category. One thousand samples, approximating the amount of subject-independent data, were generated for each subject's test. The average classification accuracy of LOO, adapting 50 real samples, adapting 1000 fake samples was 75.71$\pm$10.38\%, 75.71$\pm$12.70\% and 79.05$\pm$10.47\%, respectively. Interestingly, the influence of individual differences in this dataset is not dominant, and the classifier could perceive the information related to separating two categories. In this case, our strategy still achieves significant improvement of 3.95\% (LOO: \textit{p-value}\ \textless\ 0.01; adapting 50 real samples: \textit{p-value}\ \textless\ 0.05). Note that the classification method was not optimized for this dataset, so the results may not be the highest.

Overall, the generalization test shows that the proposed method is not data-dependent. And it has better performance for some complex data with larger individual difference.

\section{Discussion}
The cross-subject problem restricts the application of EEG-based BCI for a long time. There are two stumbling blocks. One is the large individual difference that makes it difficult to classify EEG with just subject-independent data. The other is the low signal-to-noise ratio, with which existing classification methods cannot easily perceive the difference between different categories. 

In this article, we propose a novel strategy to solve this problem from two angles, data augmentation and feature enhancement, on the basis of adaptive training. A generative adversarial network named CS-GAN has been developed to augment a very small amount of subject-specific data for adaptively training the classifier with subject-independent data. By this way, the calibration time before the actual use of a BCI system could be significantly reduced, and the classifier is also well trained with enough subject-specific information. 

Many related studies implement algorithms source from the field of image processing but ignore the inherent spatial characteristics of EEG signals, which is exactly what this paper focuses. We use the covariance matrix to construct a subspace, in which the discrimination of different categories is enlarged a lot. The spatial information obtained in the process is further used as constraints, CS-module, cov-loss and ev-loss, in CS-GAN to generate high-quality multi-channel data with implicit discrimination. Even some good research using generative models only generate EEG with few channels or just some specific features. Through this augmentation strategy and CS-GAN, we have greatly improved the performance of cross-subject EEG classification. Moreover, we design a CNN for motor imagery EEG classification, along with the idea of spatial enhancement. Remarkable results have been achieved on a general public dataset.   

It is worth noting that we also give a potential way of applying GANs, which often appears in some entertainment scenes. Although GANs-based data augmentation has been proposed for a long time, there is a paradox that it is impossible to introduce new information for deeper learning with the original dataset. Nevertheless, we use it to introduce individual information into the total dataset and prove that it is more effective than other augmentation methods. 

This paper also has some limitations and issues that need further verification. We focus on studied cross-subject, but no cross-session. The difference in the performance of different sessions is also not small. That is because we think cross-subject is a more urgent problem, and cross-session can be alleviated by controlling the objective factors such as use environment, impedance and user mood.

Another is that actually we conduct adaptive training by mixing the subject-independent data and augmentation data to train the classifier at the same time, instead of strictly pre-training with subject-independent data and fine-tuning with augmentation data. This practice refers to \cite{[60]}, and no significant difference between the two ways has been observed in our pre-tests. 

One more is about to evaluate the quality of the generated data. There is no widely accepted method to evaluate the quality of the generated EEG signal. So we choose to compare the signal itself from the time domain, frequency domain and spatial domain. In fact, it can be seen that the distribution of the generated signals is not perfect, due to more Gaussian noise inputted to CS-GAN. However, if we use shorter input like other GANs, the generator can easily fool the discriminator that only learns a small amount of real data, so that the model collapses. That is a balance of noise and diversity that worthy more investigating.

\section{Conclusion}
This paper proposes CS-GAN for EEG signals, in which the spatial features are used to generate EEG data close to the original data distribution and increase the category discrimination. Experimental results show that the cross-subject problem in EEG-based BCI has been successfully alleviated using the generated data. Besides, we also give a good classification method that can be used as a benchmark. Great potential the proposed framework has to improve the practicality of BCI.

\section*{Acknowledgment}
This work was supported in part by the National Natural Science Foundation of China (Grant No. 52075177), Joint Fund of the Ministry of Education for Equipment Pre-Research (Grant No. 6141A02033124), Research Foundation of Guangdong Province (Grant No. 2019A050505001 and 2018KZDXM002), and Guangzhou Research Foundation (Grant No. 202002030324 and 201903010028).

\ifCLASSOPTIONcaptionsoff
  \newpage
\fi

\bibliographystyle{IEEEtran}
\bibliography{ref}
\end{document}